\documentclass[11pt]{article}

\usepackage[preprint]{acl}

\usepackage{microtype}
\usepackage{booktabs}
\usepackage{array}
\usepackage{graphicx}
\usepackage{enumitem}
\usepackage{xcolor}
\usepackage{colortbl}
\setlist{nosep, leftmargin=*}

\usepackage[english,bidi=default]{babel}
\babelfont{rm}[Path=fonts/, Extension=.otf, UprightFont=*-regular, BoldFont=*-bold,
  ItalicFont=*-italic, BoldItalicFont=*-bolditalic]{texgyretermes}
\babelprovide[import]{arabic}
\babelfont[*arabic]{rm}[Path=fonts/, Extension=.ttf, UprightFont=*-Regular, BoldFont=*-Bold]{NotoNaskhArabic}

\newcommand{\sd}[1]{\foreignlanguage{arabic}{#1}}      
\newcommand{\code}[1]{\texttt{\detokenize{#1}}}         
\definecolor{oursblue}{RGB}{220,230,241}

\title{A Corpus-Driven SentencePiece Unigram Tokenizer for Sindhi:\\
Design and Benchmark Against Fourteen Existing Tokenizers}

\author{Ali Nawaz \and Muqadas Fareed \and Soonh Sadaf \\
  Department of Computer Science, Shaikh Ayaz University, Shikarpur, Pakistan \\
  \texttt{alinawaz.code@gmail.com} (corresponding author) \\[0.6em]
  Supervisor: Dr.\ Attaullah Sahito, Head of the Department of Computer Science}

\begin{document}
\maketitle

\begin{abstract}
Sindhi is a low-resource Indo-Aryan language written mainly in an extended Perso-Arabic script. Several Sindhi tokenizers have been released on public model hubs, but they have not been compared with one another on shared data. We train SentencePiece Unigram tokenizers with byte fallback at four vocabulary sizes (32k to 150k) on about 1.10 million Sindhi sentences and a smaller, repeated Urdu, Persian and Arabic slice, from four public sources, and compare them with fourteen existing tokenizers on four evaluation sets: FLORES+, Sindhi Wikipedia, a news-derived sentiment corpus, and a contemporary news corpus. The metrics are fertility, single-token retention, UNK rate, byte-fallback rate, and round-trip character loss. At 64k, our tokenizer averages 1.147 tokens per word, against 1.208 for the best existing tokenizer and 1.512 for XLM-RoBERTa, and keeps 88.6\% of words as single tokens. On fertility, single-token retention and UNK rate it is best or tied for best in all four evaluation sets at 64k and above; at 32k it trails one baseline on FLORES+. Margins over the best Sindhi-adapted baselines are small (0.002 fertility on FLORES+ at 64k), and we report no significance tests. A round-trip check finds no character loss for our tokenizer and silent character deletion (up to 0.85\% of characters) in two existing ones; seven others lose less than 0.1\%. We evaluate intrinsic metrics only, so effects on downstream tasks remain untested. Code and tokenizer files are public (DOI: 10.5281/zenodo.23962474).
\end{abstract}

\section{Introduction}
\label{sec:intro}

Sindhi is an Indo-Aryan language spoken in the Sindh province of Pakistan and in parts of India, with about 33.5 million speakers in Sindh province alone in the 2023 Pakistani census \citep{tribune2024}. It is written mainly in an extended Perso-Arabic script. Sindhi is a morphologically rich language with few public text resources \citep{ali2019corpus}, and word segmentation is itself difficult because of space omission and space insertion \citep{ali2020segmentation}.

Tokenization is the first step of every language model. It sets the sequence length for a given text, and with it the cost and the usable context, and the gap can be large between languages \citep{petrov2023}. Tokenizer design also affects downstream accuracy \citep{rust2021,schmidt2024}. A tokenizer trained mostly on other languages tends to split text in a low-resource language into many pieces, and byte-level models can need several times as many tokens for some languages \citep{petrov2023}.

Several Sindhi tokenizers have been released on public model hubs. Each is documented on its own page with its own test text, and we found no study that compares them (the search is described in Section~\ref{sec:baselines}, and we do not claim it is exhaustive). This paper makes four contributions:
\begin{enumerate}
  \item A comparison of fourteen existing tokenizers on four evaluation sets with one set of metrics, and a documented search protocol for finding them (Section~\ref{sec:baselines}).
  \item A SentencePiece Unigram tokenizer with byte fallback at four vocabulary sizes, released with training code, corpus provenance and license notes \citep{nawaz2026tokenizer} (Sections~\ref{sec:ours} and~\ref{sec:data}).
  \item A round-trip check that counts characters deleted by encoding and decoding, applied to every tokenizer. It finds silent deletion in two existing tokenizers that standard metrics do not flag (Section~\ref{sec:roundtrip}).
  \item Two design choices that did not help in our setting (clitic pre-marking and forcing frequent words to single tokens), and a record of the corrections we made to our own evaluation code (Sections~\ref{sec:design} and~\ref{sec:corrections}).
\end{enumerate}
All results are intrinsic. We train no language models and evaluate no downstream tasks, and the margins over the closest competitors are small. Section~\ref{sec:limitations} lists the limitations.

\section{Background: Sindhi Script and Text}
\label{sec:background}

Sindhi uses letters that have no counterpart in the Arabic or Persian alphabets: implosives (\sd{ٻ ڄ ڏ ڳ}), aspirated consonants written with single letters (\sd{ڀ ٿ ڇ ٺ ڦ}), and the retroflex nasal \sd{ڻ}. Urdu writes aspirated consonants as digraphs rather than single letters. A tokenizer that lacks these letters must split them into bytes or drop them.

Text in the wild mixes code points for visually similar letters, for example U+0647 and U+06C1 for heh, and some text uses Arabic Presentation Forms. SentencePiece applies Unicode normalization before training; we use its default rule, \code{nmt_nfkc}, and apply no Sindhi-specific normalization. We treat all code points that occur in the evaluation text as input to be preserved, and we take no position on their status in standard orthography.

Word boundaries are unreliable in Sindhi, so word-based metrics are noisy \citep{ali2020segmentation}. Our fertility and single-token retention use whitespace-delimited words, and they inherit this noise. This study covers Arabic-script Sindhi only. Sindhi is also written in Devanagari in India, and we do not evaluate it.

\section{Related Work}
\label{sec:related}

\subsection{Subword tokenization}

Byte-pair encoding (BPE) builds a vocabulary by repeatedly merging the most frequent symbol pair \citep{sennrich2016}. The Unigram language model instead starts from a large candidate vocabulary and keeps the pieces that maximize the likelihood of the training text \citep{kudo2018subword}. SentencePiece implements both algorithms, treats text as a raw character stream, and is designed so that decoding the encoded normalized text returns the normalized text \citep{kudo2018sentencepiece}. \citet{bostrom2020} report that Unigram tokenization recovers units closer to morphology than BPE and matches or outperforms BPE in pretraining experiments on English and Japanese. We use Unigram for Sindhi without testing it against BPE (Section~\ref{sec:limitations}).

\subsection{Evaluating tokenizers}

\citet{rust2021} show that the tokenizer is a major factor in the monolingual performance of multilingual models. \citet{petrov2023} measure tokenization parity on parallel FLORES text and find large differences in encoded length across languages. \citet{schmidt2024} test the hypothesis that fewer tokens give better downstream performance and find that it does not hold in general. Fertility and compression are therefore proxies, and we treat them as such.

\subsection{Tokenization for Indic languages and Sindhi}

\citet{pattnayak2025} compare BPE, SentencePiece and character-level tokenization for named entity recognition in low-resource Indic languages, and report results for Sindhi, which is written in Arabic script. Work on Sindhi text resources has built a web corpus of more than 61 million words for word embeddings \citep{ali2019corpus} and a neural word segmenter \citep{ali2020segmentation}. We do not know of a published comparison of Sindhi tokenizers on shared evaluation text. The tokenizers we compare (Section~\ref{sec:baselines}) are distributed as model-hub repositories without accompanying papers.

\section{Baseline Tokenizers}
\label{sec:baselines}

We evaluate fourteen existing tokenizers in three groups (Table~\ref{tab:baselines}). The first group has four Sindhi-specific tokenizers found at the start of the project: three released under the name SindhiLM (v1 to v3, differing in algorithm and vocabulary size) and one independent Sindhi BPE tokenizer. The second has six general-purpose multilingual tokenizers that were not trained with Sindhi as a primary target. The third has four Sindhi-adapted tokenizers found later: mBERT extended with a 20K-token Sindhi Unigram vocabulary, TinyLlama extended with 10,681 Sindhi tokens, and two tokenizers from the hellosindh organization.

\begin{table*}[t]
\centering
\small
\renewcommand{\arraystretch}{1.0}
\setlength{\tabcolsep}{4pt}
\begin{tabular}{llrl}
\toprule
\textbf{Tokenizer} & \textbf{Group} & \textbf{Vocabulary} & \textbf{Reference} \\
\midrule
SindhiLM v1 & Sindhi-specific (initial) & 43 & \citep{sindhilm1} \\
SindhiLM v2 & Sindhi-specific (initial) & 151,643 & \citep{sindhilm2} \\
SindhiLM v3 & Sindhi-specific (initial) & 32,000 & \citep{sindhilm3} \\
FahadQazi BPE & Sindhi-specific (initial) & 13,031 & \citep{fahadqazi} \\
XLM-RoBERTa & General multilingual & 250,002 & \citep{conneau2020} \\
mT5 & General multilingual & 250,100 & \citep{xue2021} \\
Qwen2.5 & General multilingual & 151,643 & \citep{qwen2024} \\
BLOOM & General multilingual & 250,680 & \citep{bloom2022} \\
Gemma 3 & General multilingual & 262,144 & \citep{gemma2025} \\
Llama 3.2 & General multilingual & 128,000 & \citep{llama3herd} \\
mBERT-Sindhi & Sindhi-adapted (later search) & 119,547 & \citep{kashif786} \\
TinyLlama-Sindhi & Sindhi-adapted (later search) & 42,681 & \citep{jamali} \\
hellosindh-BERT & Sindhi-adapted (later search) & 32,000 & \citep{hellosindhbert} \\
hellosindh-NER & Sindhi-adapted (later search) & 32,001 & \citep{hellosindhner} \\
\bottomrule
\end{tabular}
\caption{Baseline tokenizers. Vocabulary sizes are as loaded by the evaluation code. The Llama~3.2 vocabulary is 128,000 regular entries, as reported by the loaded tokenizer.}
\label{tab:baselines}
\end{table*}

\paragraph{Search protocol.} We searched the Hugging Face Hub and GitHub with ten search terms (``Sindhi Tokenizer'', ``Sindhi NLP'', ``Sindhi Subword Tokenization'', ``Sindhi Arabic Script'', ``Sindhi BPE'', ``Sindhi SentencePiece'', ``Sindhi Language Model'', ``Sindhi Tokenization Benchmark'', ``Low Resource Language Tokenizer'' and ``Indic Language Tokenization'') for public tokenizer files that load without modification and are described as supporting Sindhi. The first search, on 15 September 2026, found ten tokenizers. Four more were found after the first benchmark had been run, and all results were recomputed with the full set. A final search before submission found no further tokenizers on 27 September 2026. We loaded every tokenizer from its own repository between 15 and 27 September 2026 without pinning a revision. The benchmark notebook in the repository records each tokenizer's repository ID and a hash of its vocabulary.

SindhiLM v2 is a re-published copy of the Qwen2.5 tokenizer: fertility, single-token retention and total token counts are identical to three decimal places in every evaluation set. We keep it in the tables as one row, ``Qwen2.5 / SindhiLM v2'', so the comparison has 13 distinct baseline rows for 14 tokenizers. SindhiLM v1 has a vocabulary of 43 entries. The two hellosindh tokenizers differ in vocabulary size, file format and tokenization behavior on identical input, so we treat them as distinct.

\section{Our Tokenizer}
\label{sec:ours}

\subsection{Training corpus}
\label{sec:corpus}

The corpus has 1,455,915 lines and about 122.4 million characters from four sources (Table~\ref{tab:corpus}). The Sindhi side has 1,098,119 sentences and 102.7 million characters. The Urdu, Persian and Arabic side has 89,449 unique sentences and 4.9 million characters; we repeated it four times so that it makes up 16.1\% of the characters, which means that about 358,000 of the 1.46 million lines are repeats. We chose sources that do not overlap with the corpus that the SindhiLM model card names as its training data, the Sindhi Mega Corpus of ambile-official \citep{megacorpus}, so that our results would not reproduce theirs by construction. The separation is not complete: that corpus lists Sindhi Wikipedia among its seven sources, and so do we. We could not inspect it, because its dataset page showed no files when we visited it. Text was split into sentences at the Arabic full stop, period, exclamation mark, Arabic question mark and line breaks, and fragments of three characters or fewer were dropped. We applied no word-count limit and no language filter. A diagnostic on the shaikhsalman sample found that 11.3\% of its sentences were mostly Latin script. We did not remove them, so the corpus contains some English text (Section~\ref{sec:limitations}).

\begin{table*}[t]
\centering
\small
\renewcommand{\arraystretch}{1.0}
\setlength{\tabcolsep}{4pt}
\begin{tabular}{>{\raggedright\arraybackslash}p{0.27\textwidth}>{\raggedright\arraybackslash}p{0.52\textwidth}>{\raggedright\arraybackslash}p{0.13\textwidth}}
\toprule
\textbf{Source} & \textbf{Content used} & \textbf{License} \\
\midrule
Sindhi Wikipedia \citep{wikimedia} & 6,000 articles (articles 2,001 to 8,000 of a seeded shuffle; the first 2,000 are held out for evaluation) & CC BY-SA \\
Sindhi-NLP \citep{sindhinlp} & Two Twitter accounts, Awami Awaz articles, Daily Kawish articles & MIT (repository) \\
shaikhsalman/sindhi-llm-corpus \citep{shaikhsalman} & 20,000-row sample (seed 7), no language filter; about 57\% of the Sindhi-side sentences & None stated \\
Urdu, Persian and Arabic Wikipedia \citep{wikimedia} & 1,500 articles each, for shared script and loanwords; repeated four times & CC BY-SA \\
\bottomrule
\end{tabular}
\caption{Training corpus. After repetition, the cross-lingual slice makes up 16.1\% of the corpus by character count.}
\label{tab:corpus}
\end{table*}

\subsection{Training}
\label{sec:training}

We train with SentencePiece \citep{kudo2018sentencepiece} using the Unigram model type, full character coverage and byte fallback (Table~\ref{tab:settings}). Four vocabulary sizes were trained: 32,000, 64,000, 100,000 and 150,000. Byte fallback means that no character is mapped to the unknown token, and we use full character coverage so that every character seen in training, including the Sindhi letters listed in Section~\ref{sec:background}, has its own piece. The 64,000 model is the default (Section~\ref{sec:discussion}).

\begin{table*}[t]
\centering
\small
\begin{tabular}{>{\raggedright\arraybackslash}p{0.30\textwidth}>{\raggedright\arraybackslash}p{0.60\textwidth}}
\toprule
\textbf{Setting} & \textbf{Value} \\
\midrule
\code{model_type} & unigram \\
\code{vocab_size} & 32,000; 64,000 (default); 100,000; 150,000 \\
\code{character_coverage} & 1.0 \\
\code{byte_fallback} & true \\
\code{split_by_whitespace} & true \\
\code{unk_id}, \code{bos_id}, \code{eos_id}, \code{pad_id} & 0, 1, 2, disabled (−1) \\
\code{normalization_rule_name} & \code{nmt_nfkc} (SentencePiece default) \\
Seed, input sentence sampling, library versions & No seed set; \code{input_sentence_size} at its default, so all sentences are used; library versions not recorded (Kaggle default image) \\
\bottomrule
\end{tabular}
\caption{Training settings (\code{notebooks/training.ipynb} in the repository; \code{training/train_tokenizer.py} holds the same settings).}
\label{tab:settings}
\end{table*}

\subsection{Design choices tested and dropped}
\label{sec:design}

Two choices motivated by work on clitic-heavy languages were tested and removed. Each is a single experiment on one corpus, so we report them as what happened in our setting, not as general findings.

\emph{Clitic pre-marking.} We separated six common postpositions and clitics with spaces before training, guided by a corpus-wide diagnostic of fused forms. For three of the six, 17\% to 39\% of occurrences looked fused, but a control character with no plausible attachment showed 0.2\%, and the apparent fusion was overlap with unrelated words. We removed the preprocessing step.

\emph{Forced single tokens.} We forced the same six clitics and two frequent loanwords to be single tokens with \code{user_defined_symbols}. The final Unigram training objective at 32k was about 14.9 with forcing and about 9.6 without, and the targeted words needed more tokens in total. We removed forcing, and all target words became single tokens without it. Objective values from runs with different constraints are not strictly comparable, so this result supports only the weaker statement that forcing did not help. Held-out fertility for both runs would be a better comparison, and we did not run it.

\section{Evaluation Setup}
\label{sec:eval}

\subsection{Evaluation sets}
\label{sec:evalsets}

We evaluate on four sets (Table~\ref{tab:evalsets}), all in full with no sampling. FLORES+ \citep{nllb2022} is a translated benchmark. The Sentiment set was built from Kawish and Awami Awaz newspaper corpora \citep{sentimentds}, so we call it news-derived, and we built it ourselves. The Sindh Line corpus \citep{sindhline} was downloaded manually under its CC BY-SA 4.0 license because it has no scriptable interface.

\begin{table*}[t]
\centering
\small
\renewcommand{\arraystretch}{1.0}
\setlength{\tabcolsep}{4pt}
\begin{tabular}{l>{\raggedright\arraybackslash}p{0.40\textwidth}rrl}
\toprule
\textbf{Set} & \textbf{Source} & \textbf{Sentences} & \textbf{Words} & \textbf{Type} \\
\midrule
FLORES+ & snd\_Arab, devtest \citep{nllb2022} & 1,012 & 25,356 & Translated benchmark \\
Wikipedia & 20231101.sd, first 2,000 articles, 3 to 100 words \citep{wikimedia} & 25,480 & 435,947 & Encyclopedic \\
Sentiment & Sindhi\_Sentiment\_dataset \citep{sentimentds} & 3,536 & 47,156 & News-derived \\
Sindh Line & Sindh Line Publishers \citep{sindhline} & 87,158 & 1,014,998 & News \\
\bottomrule
\end{tabular}
\caption{Evaluation sets.}
\label{tab:evalsets}
\end{table*}

The Wikipedia set is disjoint from training by article. The Sentiment set and the Sindhi-NLP part of the training corpus draw on the same newspapers (Awami Awaz and Daily Kawish). The specific files do not overlap, but we did not run a sentence-level near-duplicate check, so results on this set may favor our tokenizer. For the Sindh Line set we found no overlap in sources, and we did not run a sentence-level check either. Removing the Sentiment set does not change the 64k result (Section~\ref{sec:main}).

\subsection{Metrics}
\label{sec:metrics}

\emph{Fertility} is the number of tokens divided by the number of whitespace-delimited words. Each word is encoded on its own, without special tokens. \emph{Single-token retention} is the percentage of words encoded as one token. \emph{UNK rate} is the percentage of output tokens equal to the tokenizer's unknown token. \emph{Byte-fallback rate} is the percentage of output tokens whose standalone decoding contains the replacement character U+FFFD. We report it as a diagnostic and not as a score, because a low value can mean that a tokenizer drops characters (Section~\ref{sec:roundtrip}).

\emph{Round-trip character loss} encodes and decodes each sentence, then normalizes the original and the decoded text with NFKC and removes zero-width characters, bidirectional control characters and all whitespace, so that routine normalization is not counted. Loss is the sum of positive length differences divided by the number of characters in the originals. A sentence is ``affected'' if its decoded text is shorter. The measure counts net deletions only, so a substituted or inserted character is not detected.

Encoding each word on its own differs from encoding it in context for byte-level BPE tokenizers that mark a leading space, and it can add a standalone word-boundary piece for some SentencePiece vocabularies. We did not correct for either. Every tokenizer was run with its own default normalization.

\subsection{Aggregation}

The sets differ in size by a factor of 86 (1,012 to 87,158 sentences). The main table uses an unweighted mean over the four sets. A pooled mean would be dominated by the Sindh Line set. Per-set values are in Appendix~\ref{app:a}.

\subsection{Corrections to the evaluation code}
\label{sec:corrections}

We corrected two problems in our own code. First, byte-fallback was first detected by matching byte-piece strings of the form \code{<0xNN>}. That method reported 0\% for several tokenizers whose rate was between 1\% and 30\%, including XLM-RoBERTa, which changed from 0.000\% to above 1\%. We replaced it with the standalone-decode check in Section~\ref{sec:metrics}, and excluded tokens that decode to an empty string (such as the lone SentencePiece word-boundary piece), because counting them inflated the rates of SentencePiece tokenizers. Second, after the four later tokenizers were added, we re-ran everything. Two words in the word-level test (Section~\ref{sec:wordlevel}) that had been outright wins for our tokenizer became ties. Other corrections made during development, logged in \code{docs/explainer.md} in the repository, include a tokenizer loader that silently loaded only three pieces, a training run in a forked process that silently produced empty models, an unencoded space in a download URL that had silently left out about 9,500 Sindhi-NLP articles, and two rewrites of the round-trip check to stop it counting normalization and whitespace changes as losses.

\section{Results}
\label{sec:results}

\subsection{Main results}
\label{sec:main}

Table~\ref{tab:main} gives the mean over the four sets, sorted by fertility. At 64k and above, our tokenizer has the lowest fertility, the highest single-token retention and the lowest UNK rate (tied at zero with several others) in every one of the 12 cells formed by three metrics and four sets. No baseline is strictly better in any of those cells. The same holds in the 9 cells that remain when the Sentiment set is removed. At 32k, our tokenizer is best or tied in 10 of 12 cells and is behind TinyLlama-Sindhi on FLORES+ fertility (1.233 against 1.193) and single-token retention (82.4\% against 84.4\%). We checked whether TinyLlama-Sindhi's deletion of one character (Section~\ref{sec:roundtrip}) explains this by normalizing the deleted character and re-measuring. Fertility changed by less than 0.002, so the 32k result stands.

\begin{table*}[t]
\centering
\small
\renewcommand{\arraystretch}{1.0}
\setlength{\tabcolsep}{4pt}
\begin{tabular}{lrrrrrr}
\toprule
\textbf{Tokenizer} & \textbf{Vocab} & \textbf{Fertility} & \textbf{Single-token \%} & \textbf{UNK \%} & \textbf{Byte-fallback \%} & \textbf{Max loss \%} \\
\midrule
\rowcolor{oursblue} Ours 150k & 150,000 & 1.125 & 89.7 & 0.000 & 0.154 & 0.0000 \\
\rowcolor{oursblue} Ours 100k & 100,000 & 1.133 & 89.3 & 0.000 & 0.149 & 0.0000 \\
\rowcolor{oursblue} Ours 64k & 64,000 & 1.147 & 88.6 & 0.000 & 0.138 & 0.0000 \\
\rowcolor{oursblue} Ours 32k & 32,000 & 1.184 & 86.5 & 0.000 & 0.113 & 0.0000 \\
TinyLlama-Sindhi\textsuperscript{\textdagger} & 42,681 & 1.208 & 84.4 & 0.000 & 0.000 & 0.4147 \\
hellosindh-NER & 32,001 & 1.218 & 83.7 & 0.194 & 0.197 & 0.0751 \\
mBERT-Sindhi & 119,547 & 1.278 & 78.4 & 0.143 & 0.000 & 0.0021 \\
XLM-RoBERTa & 250,002 & 1.512 & 65.4 & 0.034 & 1.402 & 0.0040 \\
SindhiLM v3 & 32,000 & 1.730 & 55.8 & 3.605 & 0.000 & 0.0754 \\
FahadQazi BPE\textsuperscript{\textdagger} & 13,031 & 1.919 & 39.1 & 0.000 & 7.127 & 0.8461 \\
hellosindh-BERT & 32,000 & 1.960 & 35.5 & 0.122 & 0.114 & 0.0089 \\
mT5 & 250,100 & 2.280 & 21.7 & 0.022 & 12.746 & 0.0040 \\
Gemma 3 & 262,144 & 2.308 & 26.8 & 0.000 & 0.105 & 0.0000 \\
Qwen2.5 / SindhiLM v2 & 151,643 & 2.580 & 23.8 & 0.000 & 8.665 & 0.0000 \\
BLOOM & 250,680 & 2.627 & 24.7 & 0.000 & 31.660 & 0.0000 \\
Llama 3.2 & 128,000 & 2.896 & 18.2 & 0.000 & 28.472 & 0.0000 \\
SindhiLM v1 & 43 & 4.306 & 0.1 & 14.017 & 0.000 & 0.0790 \\
\bottomrule
\end{tabular}
\caption{Mean over the four evaluation sets, sorted by fertility (lower is better); single-token retention is higher-is-better. Byte-fallback is a diagnostic. ``Max loss'' is the largest round-trip character loss in any set (Table~\ref{tab:roundtrip}). Rows for our tokenizer are shaded. \textsuperscript{\textdagger}~Zero UNK rate comes with silent character deletion (Section~\ref{sec:roundtrip}).}
\label{tab:main}
\end{table*}

At 64k, mean fertility is 24.1\% lower than XLM-RoBERTa (1.147 against 1.512) and 60.4\% lower than Llama~3.2 (2.896). Those tokenizers were not built for Sindhi. The relevant comparison is with the Sindhi-adapted tokenizers. Against the nearest one, TinyLlama-Sindhi, the mean reduction is 5.0\% (0.061 tokens per word). Table~\ref{tab:margins} shows the margin in each set.

\begin{table*}[t]
\centering
\footnotesize
\renewcommand{\arraystretch}{1.0}
\setlength{\tabcolsep}{4pt}
\begin{tabular}{lcccccc}
\toprule
\textbf{Set} & \textbf{Fertility: Ours} & \textbf{Best baseline} & \textbf{Diff.} & \textbf{Single-token \%: Ours} & \textbf{Best baseline} & \textbf{Diff.} \\
\midrule
FLORES+ & 1.191 & 1.193 (TinyLlama-Sindhi) & 0.002 & 84.7 & 84.4 (TinyLlama-Sindhi) & +0.3 \\
Wikipedia & 1.194 & 1.272 (hellosindh-NER) & 0.078 & 86.9 & 82.6 (hellosindh-NER) & +4.3 \\
Sentiment & 1.110 & 1.149 (TinyLlama-Sindhi) & 0.039 & 90.3 & 87.6 (TinyLlama-Sindhi) & +2.7 \\
Sindh Line & 1.093 & 1.155 (hellosindh-NER) & 0.062 & 92.4 & 87.9 (hellosindh-NER) & +4.5 \\
Mean & 1.147 & 1.208 (TinyLlama-Sindhi) & 0.061 & 88.6 & 84.4 (TinyLlama-Sindhi) & +4.2 \\
\bottomrule
\end{tabular}
\caption{Margin of Ours 64k over the best baseline in each set. Differences are absolute (fertility in tokens per word, single-token retention in percentage points). No confidence intervals are available.}
\label{tab:margins}
\end{table*}

On FLORES+ the margin at 64k is 0.002 fertility and 0.3 points of single-token retention. Without confidence intervals we treat this as parity. On the other three sets the margin is 0.039 to 0.078 tokens per word.

Vocabulary size confounds fertility, since a larger vocabulary generally lowers it. Table~\ref{tab:matched} compares tokenizers with about 32,000 entries. Ours at 32k has lower fertility than all three (1.184 against 1.218 to 1.960). The strongest baselines mostly have larger vocabularies (TinyLlama-Sindhi 42,681; mBERT-Sindhi 119,547), and Ours 32k still has lower mean fertility than TinyLlama-Sindhi.

\begin{table*}[t]
\centering
\small
\renewcommand{\arraystretch}{1.0}
\setlength{\tabcolsep}{4pt}
\begin{tabular}{lrrrr}
\toprule
\textbf{Tokenizer} & \textbf{Vocab} & \textbf{Fertility} & \textbf{Single-token \%} & \textbf{UNK \%} \\
\midrule
\rowcolor{oursblue} Ours 32k & 32,000 & 1.184 & 86.5 & 0.000 \\
hellosindh-NER & 32,001 & 1.218 & 83.7 & 0.194 \\
SindhiLM v3 & 32,000 & 1.730 & 55.8 & 3.605 \\
hellosindh-BERT & 32,000 & 1.960 & 35.5 & 0.122 \\
TinyLlama-Sindhi (larger vocabulary) & 42,681 & 1.208 & 84.4 & 0.000 \\
\bottomrule
\end{tabular}
\caption{Tokenizers with about 32,000 entries, with TinyLlama-Sindhi as a larger-vocabulary reference. Means over four sets.}
\label{tab:matched}
\end{table*}

\begin{figure*}[t]
  \centering
  \includegraphics[width=0.72\textwidth]{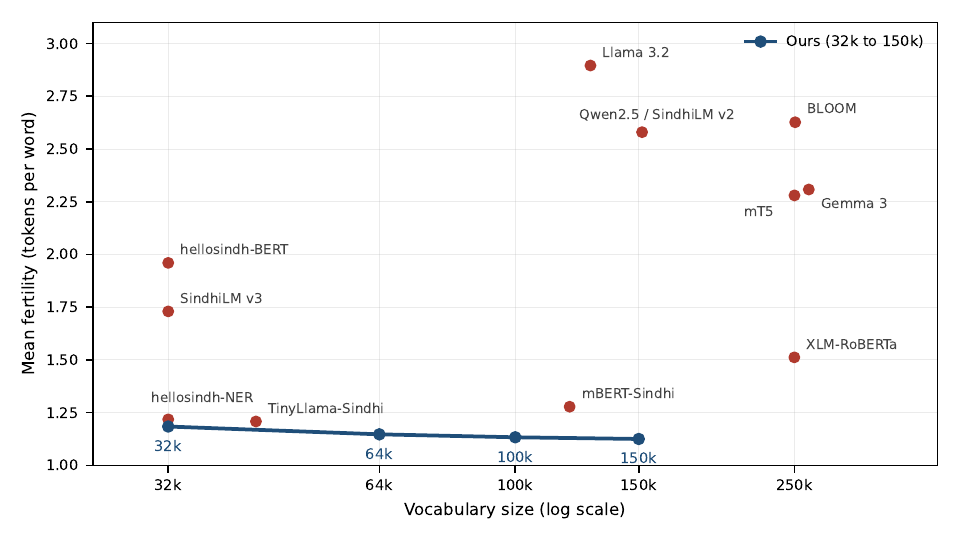}
  \caption{Mean fertility against vocabulary size (log scale). Our tokenizers (line) lie below the baselines at every size. SindhiLM v1 (vocabulary 43, fertility 4.306) is omitted.}
  \label{fig:vocab}
\end{figure*}

Gains diminish with vocabulary size: mean fertility is 1.184, 1.147, 1.133 and 1.125 at 32k, 64k, 100k and 150k. Going from 64k to 150k lowers fertility by 1.9\%. The 64k model is the smallest size that is best or tied in all 12 cells, and we use it as the default. The recommendation rests on intrinsic metrics only.

\subsection{Per-set results}
\label{sec:perset}

\begin{figure*}[t]
  \centering
  \includegraphics[width=0.72\textwidth]{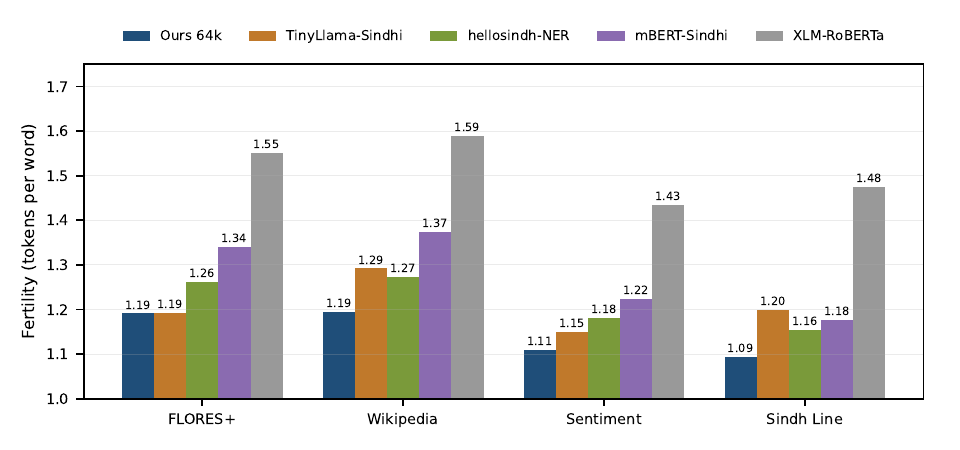}
  \caption{Fertility by evaluation set for Ours 64k and the four strongest baselines. Full per-set results for all tokenizers are in Table~\ref{tab:a1}.}
  \label{fig:perset}
\end{figure*}

Figure~\ref{fig:perset} shows fertility in each evaluation set for our 64k tokenizer and the four strongest baselines. Ours has the lowest value in all four sets; on FLORES+, TinyLlama-Sindhi is within 0.002.

\subsection{Round-trip character loss}
\label{sec:roundtrip}

TinyLlama-Sindhi and FahadQazi BPE both report a UNK rate of exactly zero, and TinyLlama-Sindhi also a byte-fallback rate of zero (Table~\ref{tab:main}), yet both delete characters. We first saw this for TinyLlama-Sindhi, when emoji, Chinese, Cyrillic and mathematical-symbol strings decoded to shorter strings with neither UNK nor byte-fallback tokens. Table~\ref{tab:roundtrip} shows the effect on real text. Five tokenizers, including all four of ours, show exactly zero loss in every set. The remaining tokenizers lose less than 0.1\% of characters, though up to 1.13\% of sentences are affected (SindhiLM v3).

\begin{table*}[t]
\centering
\small
\renewcommand{\arraystretch}{1.0}
\setlength{\tabcolsep}{4pt}
\begin{tabular}{lrlrl}
\toprule
\textbf{Tokenizer} & \textbf{Max loss \%} & \textbf{Set} & \textbf{Max sentences affected \%} & \textbf{Set} \\
\midrule
FahadQazi BPE & 0.8461 & Wikipedia & 16.15 & Wikipedia \\
TinyLlama-Sindhi & 0.4147 & FLORES+ & 24.90 & FLORES+ \\
SindhiLM v1 & 0.0790 & Wikipedia & 0.69 & Wikipedia \\
SindhiLM v3 & 0.0754 & Wikipedia & 1.13 & Wikipedia \\
hellosindh-NER & 0.0751 & Wikipedia & 0.49 & Wikipedia \\
hellosindh-BERT & 0.0089 & Wikipedia & 0.09 & Wikipedia \\
XLM-RoBERTa & 0.0040 & Wikipedia & 0.04 & Wikipedia \\
mT5 & 0.0040 & Wikipedia & 0.04 & Wikipedia \\
mBERT-Sindhi & 0.0021 & Wikipedia & 0.04 & Wikipedia \\
\bottomrule
\end{tabular}
\caption{Round-trip character loss: the largest value over the four sets, with the set where it occurs. Tokenizers with zero loss in every set: Ours (all sizes), Qwen2.5 / SindhiLM v2, BLOOM, Gemma~3, Llama~3.2. Per-set values are in Table~\ref{tab:a2}.}
\label{tab:roundtrip}
\end{table*}

FahadQazi BPE deletes individual Latin letters in code-switched text. For example, the sentence ``An update on my situation, stay safe everyone'' decodes with the letters A, p, f and v, among others, removed. This affects up to 16.15\% of Wikipedia sentences, where citations and transliterations are common. TinyLlama-Sindhi deletes every occurrence of U+06C1 (ARABIC LETTER HEH GOAL), which occurs in common function words such as \sd{تہ}, \sd{بہ} and \sd{نہ}, and it corrupts text written in the Arabic Presentation Forms block, in the worst case reducing a sentence to seven isolated characters. U+06C1 appears in 24.8\% of FLORES+ sentences and 6.2\% of Wikipedia sentences, which explains why its loss is largest on FLORES+.

Zero loss for our tokenizer is expected from the design, since byte fallback keeps every character representable, and NFKC is applied to both sides before comparison. The informative result is for the baselines. Lossless decoding is a design goal of SentencePiece \citep{kudo2018sentencepiece}. We use round-trip loss as a reported metric for comparing tokenizers, not as a unit test, and we recommend reporting it next to fertility and UNK rate.

\subsection{Word-level test}
\label{sec:wordlevel}

We tokenized 22 hand-selected words (four categories: genitive constructions, inflectional forms, aspirated digraphs, and loanwords) with every tokenizer and compared token counts (Appendix~\ref{app:b}). Our tokenizer ties the best competitor on 21 of 22 words and is behind on one (\sd{هنجي}, 2 tokens against 1 for FahadQazi BPE and mBERT-Sindhi). It wins outright on none. The list was fixed before the four later tokenizers were found and re-run unchanged. This test counts tokens only. It does not check where the boundaries fall, so it is not a morphological evaluation. It shows that several competitors are already near-optimal on common, well-behaved words.

\section{Discussion}
\label{sec:discussion}

The results support three statements. First, the new tokenizer has lower mean fertility and higher single-token retention than all fourteen existing tokenizers in this benchmark, by a large margin over general multilingual tokenizers and by a small one (5\%) over the best Sindhi-adapted tokenizer. Second, the advantage over the closest competitors is not uniform: it is 0.002 on FLORES+ and 0.04 to 0.08 tokens per word elsewhere. Third, UNK and byte-fallback rates can look excellent because a tokenizer discards what it cannot represent, and the round-trip check is what exposes this.

We do not know why our tokenizer does better. Plausible factors are a larger and more varied training corpus than the baselines used, and the Unigram algorithm, but we did not train a BPE tokenizer on the same corpus or vary corpus size, and the baselines differ in vocabulary size, algorithm and training data that is mostly undocumented. These remain hypotheses.

Lower fertility does not by itself make a better language model \citep{schmidt2024,rust2021}. A 5\% reduction in tokens per word may or may not matter downstream. The byte-fallback correction in Section~\ref{sec:corrections} is a reminder that a plausible detection heuristic can silently misreport. A heuristic that reported 0\% for XLM-RoBERTa was wrong, and the corrected value is above 1\%. We have not broken the corrected value down by piece for XLM-RoBERTa or mT5.

\section{Limitations}
\label{sec:limitations}

\begin{enumerate}
  \item \emph{Intrinsic evaluation only.} We train no language model and run no downstream task. The default vocabulary size and the ranking of tokenizers rest on proxy metrics.
  \item \emph{No significance tests.} We report no confidence intervals. The 64k margin on FLORES+ (0.002) should be read as parity.
  \item \emph{No ablation.} We did not compare Unigram with BPE on the same corpus, and did not vary corpus size, so we cannot attribute the advantage to either factor.
  \item \emph{Evaluation-set independence.} The Sentiment set shares newspaper sources with part of the training corpus and was built by the authors. We ran no sentence-level near-duplicate check on it or on the Sindh Line set.
  \item \emph{Training corpus quality.} The training text was not filtered by language or length. The shaikhsalman sample contains English text (11.3\% of its sentences were mostly Latin script in our diagnostic), the Urdu, Persian and Arabic slice is repeated four times, and Sindhi Wikipedia may also appear in the SindhiLM training corpus. We did not measure how these choices affect the results.
  \item \emph{Baseline coverage and protocol.} The search for tokenizers may be incomplete. We know of two further public Sindhi tokenizers that we did not evaluate: the Sindhi Wikipedia tokenizers in the \code{wikilangs/sd} repository (8k to 64k) and the SindhiFormer tokenizer by the author of SindhiLM. Baselines were run with default normalization and with words encoded in isolation, and their vocabulary sizes differ.
  \item \emph{Metric limits.} Whitespace-delimited words are noisy in Sindhi \citep{ali2020segmentation}. The byte-fallback detector counts any token whose standalone decoding contains U+FFFD, and we did not inspect which pieces trigger it for each tokenizer. Round-trip loss counts net deletions only. The unweighted mean over four sets is a choice, not a given.
  \item \emph{No classical or literary Sindhi.} Five candidate sources were investigated, including Bhittaipedia and the Shah Jo Risalo labeled dataset \citep{bhittaipedia}. They failed for different reasons: a corrupted CSV export, an unrelated SQL dump, two unusable OCR scans (one with an incompatible license), and a subscription-gated platform.
  \item \emph{Scope.} Arabic-script Sindhi only. Dialects are not examined, and the tokenizer is not validated for speech recognition or text-to-speech.
\end{enumerate}

\section{Data, Code and License}
\label{sec:data}

The code is released under the MIT license and the tokenizer files under CC BY-SA 4.0 \citep{nawaz2026tokenizer}. We chose CC BY-SA because part of the training data is share-alike Wikipedia text; this is a cautious default and not legal advice. The repository redistributes no training text. The shaikhsalman corpus states no license and supplied about 57\% of the Sindhi-side training sentences. The Sindhi-NLP repository is MIT-licensed, but the tweets and newspaper articles in it may carry other rights. A tokenizer vocabulary can contain frequent strings from its training text, including personal and place names. Users should check these terms for their own use. Training and evaluation sources, sizes and licenses are listed in \code{DATA_SOURCES.md} in the repository. The evaluation code, per-set results and the exact release used here are in the repository at release v1.0.0 (Zenodo DOI: \href{https://doi.org/10.5281/zenodo.23962474}{10.5281/zenodo.23962474}). All Hugging Face, GitHub and Mozilla Data Collective resources were accessed between 15 and 27 September 2026, except the Sindhi Mega Corpus \citep{megacorpus} (4 November 2026).

\paragraph{Author contributions.} A.N.: conceptualization, methodology, software, data curation, investigation, formal analysis, writing -- original draft. M.F. and S.S.: validation, visualization, writing -- review and editing. A.S.: supervision.

\paragraph{Acknowledgments.} The authors thank Dr.\ Attaullah Sahito for supervision and guidance, Kaggle for the compute used for training and evaluation, and the creators of the datasets we used: the Sindhi-NLP project, shaikhsalman, Sindh Line Publishers, Open Language Data (FLORES+) and the Wikimedia Foundation.

\paragraph{Funding.} This research received no external funding.

\paragraph{Competing interests.} The authors declare no competing interests.

\section{Conclusion}
\label{sec:conclusion}

A SentencePiece Unigram tokenizer trained on about 1.10 million Sindhi sentences and a smaller, repeated Urdu, Persian and Arabic slice, from four public sources reaches a mean fertility of 1.147 tokens per word at 64k, against 1.208 for the best of fourteen existing tokenizers. It is best or tied for best on fertility, single-token retention and UNK rate in every evaluation set at 64k and above, with a margin of 0.002 on FLORES+ and 0.04 to 0.08 elsewhere. It loses no characters in a round-trip check, while two existing tokenizers silently delete characters (up to 0.85\% of characters) and seven others lose less than 0.1\%. These are intrinsic results on four evaluation sets and carry the limits in Section~\ref{sec:limitations}. The next steps are to train small language models with the leading tokenizers on identical data and compare them, to add confidence intervals and near-duplicate checks, to run a BPE and corpus-size ablation, to measure tokenization parity against English on FLORES+, to build a small gold morpheme-segmentation set, and to add the unevaluated tokenizers and a literary evaluation set.

\bibliography{references}

\appendix
\clearpage
\onecolumn

\section{Per-Set Results}
\label{app:a}
\renewcommand{\thetable}{A\arabic{table}}
\setcounter{table}{0}

\begin{table*}[htbp]
\centering
\begin{minipage}[t]{0.49\textwidth}
\centering
\scriptsize
\begin{tabular}{lrrrr}
\toprule
\multicolumn{5}{c}{\textbf{FLORES+}} \\
\textbf{Tokenizer} & \textbf{Fert.} & \textbf{Single \%} & \textbf{UNK \%} & \textbf{Byte-fb. \%} \\
\midrule
\rowcolor{oursblue} Ours 150k & 1.167 & 85.9 & 0.000 & 0.199 \\
\rowcolor{oursblue} Ours 100k & 1.176 & 85.5 & 0.000 & 0.198 \\
\rowcolor{oursblue} Ours 64k & 1.191 & 84.7 & 0.000 & 0.172 \\
\rowcolor{oursblue} Ours 32k & 1.233 & 82.4 & 0.000 & 0.099 \\
TinyLlama-Sindhi & 1.193 & 84.4 & 0.000 & 0.000 \\
hellosindh-NER & 1.261 & 79.7 & 0.106 & 0.109 \\
mBERT-Sindhi & 1.341 & 73.9 & 0.056 & 0.000 \\
XLM-RoBERTa & 1.552 & 61.8 & 0.010 & 1.303 \\
SindhiLM v3 & 1.824 & 52.7 & 4.650 & 0.000 \\
FahadQazi BPE & 1.940 & 37.7 & 0.000 & 7.199 \\
hellosindh-BERT & 1.983 & 34.5 & 0.068 & 0.070 \\
mT5 & 2.318 & 19.9 & 0.007 & 13.040 \\
Gemma 3 & 2.341 & 25.6 & 0.000 & 0.000 \\
Qwen2.5 / SindhiLM v2 & 2.617 & 22.3 & 0.000 & 9.223 \\
BLOOM & 2.692 & 23.4 & 0.000 & 32.700 \\
Llama 3.2 & 2.951 & 17.0 & 0.000 & 30.015 \\
SindhiLM v1 & 4.307 & 0.0 & 14.019 & 0.000 \\
\bottomrule
\end{tabular}
\end{minipage}\hfill
\begin{minipage}[t]{0.49\textwidth}
\centering
\scriptsize
\begin{tabular}{lrrrr}
\toprule
\multicolumn{5}{c}{\textbf{Wikipedia}} \\
\textbf{Tokenizer} & \textbf{Fert.} & \textbf{Single \%} & \textbf{UNK \%} & \textbf{Byte-fb. \%} \\
\midrule
\rowcolor{oursblue} Ours 150k & 1.157 & 88.6 & 0.000 & 0.179 \\
\rowcolor{oursblue} Ours 100k & 1.171 & 88.0 & 0.000 & 0.173 \\
\rowcolor{oursblue} Ours 64k & 1.194 & 86.9 & 0.000 & 0.167 \\
\rowcolor{oursblue} Ours 32k & 1.249 & 84.0 & 0.000 & 0.158 \\
TinyLlama-Sindhi & 1.292 & 80.5 & 0.000 & 0.000 \\
hellosindh-NER & 1.272 & 82.6 & 0.495 & 0.411 \\
mBERT-Sindhi & 1.373 & 73.5 & 0.440 & 0.000 \\
XLM-RoBERTa & 1.589 & 63.1 & 0.111 & 1.774 \\
SindhiLM v3 & 1.726 & 55.4 & 4.949 & 0.000 \\
FahadQazi BPE & 2.002 & 37.4 & 0.000 & 8.613 \\
hellosindh-BERT & 1.988 & 36.1 & 0.317 & 0.246 \\
mT5 & 2.285 & 22.1 & 0.077 & 13.079 \\
Gemma 3 & 2.317 & 28.0 & 0.000 & 0.259 \\
Qwen2.5 / SindhiLM v2 & 2.568 & 25.5 & 0.000 & 7.498 \\
BLOOM & 2.569 & 25.3 & 0.000 & 29.959 \\
Llama 3.2 & 2.856 & 18.5 & 0.000 & 26.716 \\
SindhiLM v1 & 4.271 & 0.2 & 15.137 & 0.000 \\
\bottomrule
\end{tabular}
\end{minipage}
\vspace{1.2em}

\begin{minipage}[t]{0.49\textwidth}
\centering
\scriptsize
\begin{tabular}{lrrrr}
\toprule
\multicolumn{5}{c}{\textbf{Sentiment}} \\
\textbf{Tokenizer} & \textbf{Fert.} & \textbf{Single \%} & \textbf{UNK \%} & \textbf{Byte-fb. \%} \\
\midrule
\rowcolor{oursblue} Ours 150k & 1.099 & 91.0 & 0.000 & 0.137 \\
\rowcolor{oursblue} Ours 100k & 1.103 & 90.8 & 0.000 & 0.133 \\
\rowcolor{oursblue} Ours 64k & 1.110 & 90.3 & 0.000 & 0.126 \\
\rowcolor{oursblue} Ours 32k & 1.130 & 89.0 & 0.000 & 0.126 \\
TinyLlama-Sindhi & 1.149 & 87.6 & 0.000 & 0.000 \\
hellosindh-NER & 1.182 & 84.6 & 0.084 & 0.068 \\
mBERT-Sindhi & 1.223 & 81.2 & 0.040 & 0.000 \\
XLM-RoBERTa & 1.434 & 69.3 & 0.009 & 1.133 \\
SindhiLM v3 & 1.739 & 55.3 & 2.941 & 0.000 \\
FahadQazi BPE & 1.831 & 42.2 & 0.000 & 6.318 \\
hellosindh-BERT & 1.915 & 36.2 & 0.052 & 0.020 \\
mT5 & 2.259 & 22.4 & 0.001 & 12.120 \\
Gemma 3 & 2.276 & 27.0 & 0.000 & 0.162 \\
Qwen2.5 / SindhiLM v2 & 2.562 & 23.5 & 0.000 & 8.968 \\
BLOOM & 2.645 & 24.7 & 0.000 & 32.653 \\
Llama 3.2 & 2.907 & 18.4 & 0.000 & 29.384 \\
SindhiLM v1 & 4.258 & 0.1 & 12.542 & 0.000 \\
\bottomrule
\end{tabular}
\end{minipage}\hfill
\begin{minipage}[t]{0.49\textwidth}
\centering
\scriptsize
\begin{tabular}{lrrrr}
\toprule
\multicolumn{5}{c}{\textbf{Sindh Line}} \\
\textbf{Tokenizer} & \textbf{Fert.} & \textbf{Single \%} & \textbf{UNK \%} & \textbf{Byte-fb. \%} \\
\midrule
\rowcolor{oursblue} Ours 150k & 1.078 & 93.3 & 0.000 & 0.100 \\
\rowcolor{oursblue} Ours 100k & 1.084 & 93.0 & 0.000 & 0.091 \\
\rowcolor{oursblue} Ours 64k & 1.093 & 92.4 & 0.000 & 0.085 \\
\rowcolor{oursblue} Ours 32k & 1.124 & 90.5 & 0.000 & 0.068 \\
TinyLlama-Sindhi & 1.198 & 85.2 & 0.000 & 0.000 \\
hellosindh-NER & 1.155 & 87.9 & 0.090 & 0.201 \\
mBERT-Sindhi & 1.176 & 85.0 & 0.038 & 0.000 \\
XLM-RoBERTa & 1.475 & 67.4 & 0.007 & 1.400 \\
SindhiLM v3 & 1.630 & 59.9 & 1.881 & 0.000 \\
FahadQazi BPE & 1.903 & 39.3 & 0.000 & 6.376 \\
hellosindh-BERT & 1.953 & 35.2 & 0.053 & 0.118 \\
mT5 & 2.259 & 22.3 & 0.004 & 12.746 \\
Gemma 3 & 2.299 & 26.7 & 0.000 & 0.000 \\
Qwen2.5 / SindhiLM v2 & 2.574 & 23.9 & 0.000 & 8.969 \\
BLOOM & 2.600 & 25.5 & 0.000 & 31.328 \\
Llama 3.2 & 2.871 & 19.0 & 0.000 & 27.774 \\
SindhiLM v1 & 4.388 & 0.0 & 14.368 & 0.000 \\
\bottomrule
\end{tabular}
\end{minipage}
\caption{Per-set results for 17 tokenizer rows (four sizes of ours and 13 baseline rows, with SindhiLM~v2 merged into Qwen2.5). Fertility is tokens per word; the other columns are percentages.}
\label{tab:a1}
\end{table*}

\begin{table*}[htbp]
\centering
\scriptsize
\renewcommand{\arraystretch}{1.0}
\setlength{\tabcolsep}{3pt}
\begin{tabular}{lrrrrrrrr}
\toprule
\textbf{Tokenizer} & \textbf{FLORES+ loss} & \textbf{FLORES+ aff.} & \textbf{Wiki loss} & \textbf{Wiki aff.} & \textbf{Sent. loss} & \textbf{Sent. aff.} & \textbf{Sindh Line loss} & \textbf{Sindh Line aff.} \\
\midrule
\rowcolor{oursblue} Ours (all sizes) & 0.0000 & 0.00 & 0.0000 & 0.00 & 0.0000 & 0.00 & 0.0000 & 0.00 \\
Qwen2.5 / SindhiLM v2 & 0.0000 & 0.00 & 0.0000 & 0.00 & 0.0000 & 0.00 & 0.0000 & 0.00 \\
BLOOM & 0.0000 & 0.00 & 0.0000 & 0.00 & 0.0000 & 0.00 & 0.0000 & 0.00 \\
Gemma 3 & 0.0000 & 0.00 & 0.0000 & 0.00 & 0.0000 & 0.00 & 0.0000 & 0.00 \\
Llama 3.2 & 0.0000 & 0.00 & 0.0000 & 0.00 & 0.0000 & 0.00 & 0.0000 & 0.00 \\
XLM-RoBERTa & 0.0000 & 0.00 & 0.0040 & 0.04 & 0.0000 & 0.00 & 0.0030 & 0.02 \\
mT5 & 0.0000 & 0.00 & 0.0040 & 0.04 & 0.0000 & 0.00 & 0.0030 & 0.02 \\
mBERT-Sindhi & 0.0000 & 0.00 & 0.0021 & 0.04 & 0.0000 & 0.00 & 0.0001 & 0.00 \\
hellosindh-BERT & 0.0000 & 0.00 & 0.0089 & 0.09 & 0.0000 & 0.00 & 0.0031 & 0.02 \\
SindhiLM v1 & 0.0000 & 0.00 & 0.0790 & 0.69 & 0.0000 & 0.00 & 0.0005 & 0.01 \\
SindhiLM v3 & 0.0010 & 0.10 & 0.0754 & 1.13 & 0.0086 & 0.31 & 0.0029 & 0.04 \\
hellosindh-NER & 0.0021 & 0.10 & 0.0751 & 0.49 & 0.0017 & 0.06 & 0.0072 & 0.11 \\
TinyLlama-Sindhi & 0.4147 & 24.90 & 0.3114 & 6.95 & 0.0409 & 1.67 & 0.0413 & 1.44 \\
FahadQazi BPE & 0.4105 & 15.81 & 0.8461 & 16.15 & 0.0847 & 1.87 & 0.2272 & 6.08 \\
\bottomrule
\end{tabular}
\caption{Round-trip character loss by set. ``Loss'' is the percentage of characters lost; ``Aff.'' is the percentage of sentences with at least one lost character. Sentiment is abbreviated ``Sent.''}
\label{tab:a2}
\end{table*}

\clearpage
\section{Word-Level Test}
\label{app:b}
\renewcommand{\thetable}{B\arabic{table}}
\setcounter{table}{0}

\begin{table*}[htbp]
\centering
\small
\renewcommand{\arraystretch}{1.0}
\setlength{\tabcolsep}{4pt}
\begin{tabular}{llccc}
\toprule
\textbf{Category} & \textbf{Word} & \textbf{Best custom} & \textbf{Best competitor} & \textbf{Result} \\
\midrule
Genitive/possessive & \sd{دنيا جو} & 32k (2) & 2 (5 tokenizers) & Tie \\
Genitive/possessive & \sd{خدا جو} & 32k (2) & 2 (6 tokenizers) & Tie \\
Genitive/possessive & \sd{پاڻي جي} & 32k (2) & 2 (5 tokenizers) & Tie \\
Genitive/possessive & \sd{ماڻهن جو} & 32k (2) & 2 (5 tokenizers) & Tie \\
Inflectional suffix & \sd{ڪتابن} & 32k (1) & 1 (5 tokenizers) & Tie \\
Inflectional suffix & \sd{شهرن} & 32k (1) & 1 (3 tokenizers) & Tie \\
Inflectional suffix & \sd{توهانڪي} & 64k (1) & 1 (hellosindh-NER) & Tie \\
Inflectional suffix & \sd{منهنجو} & 32k (1) & 1 (5 tokenizers) & Tie \\
Inflectional suffix & \sd{هنجي} & 32k (2) & 1 (FahadQazi BPE, mBERT-Sindhi) & FahadQazi BPE \\
Aspirated digraph & \sd{ٻار} & 32k (1) & 1 (5 tokenizers) & Tie \\
Aspirated digraph & \sd{ڄاڻ} & 32k (1) & 1 (5 tokenizers) & Tie \\
Aspirated digraph & \sd{ڏينهن} & 32k (1) & 1 (4 tokenizers) & Tie \\
Aspirated digraph & \sd{ڳوٺ} & 32k (1) & 1 (7 tokenizers) & Tie \\
Aspirated digraph & \sd{ڀاءُ} & 32k (1) & 1 (4 tokenizers) & Tie \\
Aspirated digraph & \sd{ٿورو} & 32k (1) & 1 (5 tokenizers) & Tie \\
Aspirated digraph & \sd{ڇوڪرو} & 32k (1) & 1 (4 tokenizers) & Tie \\
Loanword & \sd{علم} & 32k (1) & 1 (8 tokenizers) & Tie \\
Loanword & \sd{تعليم} & 32k (1) & 1 (9 tokenizers) & Tie \\
Loanword & \sd{دنيا} & 32k (1) & 1 (6 tokenizers) & Tie \\
Loanword & \sd{اسڪول} & 32k (1) & 1 (5 tokenizers) & Tie \\
Loanword & \sd{ڪمپيوٽر} & 32k (1) & 1 (5 tokenizers) & Tie \\
Loanword & \sd{انٽرنيٽ} & 32k (1) & 1 (5 tokenizers) & Tie \\
\bottomrule
\end{tabular}
\caption{Token counts for 22 hand-selected words. ``Best custom'' is the best of our four sizes (token count in brackets). ``Best competitor'' is the best of the 13 non-duplicate baselines (SindhiLM~v2 and Qwen2.5 are identical). The result is 21 ties, 0 outright wins and 1 loss.}
\label{tab:b1}
\end{table*}

\end{document}